\documentclass[10pt,twocolumn,letterpaper]{article}
\usepackage{iccv}
\usepackage{times}
\usepackage{epsfig}
\usepackage{graphicx}
\usepackage{amsmath}
\usepackage{amssymb}
\usepackage{algorithm} 
\usepackage{algorithmic}
\usepackage{multirow}
\usepackage{cite}

\usepackage[breaklinks=true,bookmarks=false]{hyperref}

\iccvfinalcopy 

\ificcvfinal\fi

\begin{document}

\title{Aggregative Self-Supervised Feature Learning from a Limited Sample}

\author{Jiuwen Zhu\\
Institute of Computing Technology, \\Chinese Academy of Sciences\\
{\tt\small zhujiuwen19g@ict.ac.cn}
\and
Yuexiang Li\\
Jarvis Lab, Tencent\\
{\tt\small vicyxli@tencent.com}
\and
S. Kevin Zhou\\
Institute of Computing Technology, Chinese Academy of Sciences\\
{\tt\small s.kevin.zhou@gmail.com}
}

\maketitle
\ificcvfinal\thispagestyle{empty}\fi

\begin{abstract}
Self-supervised learning (SSL) is an efficient approach that addresses the issue of limited training data and annotation shortage. The key part in SSL is its proxy task that defines the supervisory signals and drives the learning toward effective feature representations. However, most SSL approaches usually focus on a single proxy task, which greatly limits the expressive power of the learned features and therefore deteriorates the network generalization capacity. In this regard, we hereby propose two strategies of \textbf{aggregation} in terms of complementarity of various forms to boost the robustness of self-supervised learned features. We firstly propose a \textbf{principled framework} of multi-task aggregative self-supervised learning from a \textbf{limited sample} to form a unified representation, with an intent of exploiting feature complementarity among different tasks. Then, in self-aggregative SSL, we propose to \textbf{self-complement} an existing proxy task with an auxiliary loss function based on a linear centered kernel alignment metric, which explicitly promotes the exploring of where are uncovered by the features learned from a proxy task at hand to further boost the modeling capability. Our extensive experiments on 2D natural image and 3D medical image classification tasks under limited data and annotation scenarios confirm that the proposed aggregation strategies successfully boost the classification accuracy.
\end{abstract}


\section{Introduction}
Recently, self-supervised learning (SSL)~\cite{Wang2015UnsupervisedLO, Owens2016AmbientSP, Chen2020BigSM, Ji2019IIC, Kolesnikov2019RevisitingSV, Jakab2020SelfSupervisedLO, Newell2020HowUI, Lai2020MASTAM} gains increasing attentions in the community as it attempts to loose the requirement of annotated data for neural networks by exploiting the rich information contained in unlabeled data, especially in the scenario of small number of data such as medical image data~\cite{Zhou2020ARO, Tajbakhsh2020EmbracingID}. A conventional SSL approach starts with a formulated proxy task to encourage the learning of informative features from raw data. A multitude of proxy tasks, dealing with 2D natural images or 3D medical volumes, have been proposed, including grayscale image colorization \cite{larsson_colorization_2017}, images rotation \cite{gidaris2018image_rotations}, Jigsaw puzzles \cite{noroozi2016jigsaw_puzzles,Arbitrary_Jigsaw_Puzzles}, BigBiGAN~\cite{Donahue2019LargeSA}, SimCLR~\cite{chen2020SimCLR}, Rubik's cube~\cite{Zhuang_2019_MICCAI, ZHURubik}, Model Genesis~\cite{zhou2019MG}, and D2D-CNNs~\cite{Blendowski2019HowTL}.

Most SSL approaches usually focus on a single proxy task, which limits their representation capacity of the learned features, and therefore decreases their generalization. Nevertheless, there are few studies trying to exploit the potential of assembling multiple SSL tasks. 
Concretely, Doersch and Zisserman~\cite{Doersch2017MultitaskSSL} made the first attempt in the area exploring the way to combine multiple self-supervised tasks. Chen \emph{et al.}~\cite{Chen2020AdversarialRF} introduced an adversarial training strategy for the assemble of self-supervised tasks. Jenni \emph{et al.}~\cite{Jenni2020LCI} proposed a SSL proxy task, which required the network to recognize different transformations generated by multiple proxy tasks. Model Genesis~\cite{ZHOU_MG_MIA, zhou2019MG} defined a pool of image permutations for network to randomly select, and designs a unitary self-supervised learning framework to reconstruct the original image.
Those approaches demonstrate that the integration of proxy tasks improves the generalization of pre-trained networks and thus boosts the performance of subsequent target tasks.

The previous mentioned task integration approaches are usually derived from ad-hoc assumptions and \textit{there is a lack of a principled way of aggregation}. In this paper, we attempt to bridge the gap by exploiting two different forms of complementarity. 
We first systematically explore the \textit{feature complementarity} between multiple SSL approaches and propose a greedy algorithm to aggregate multiple proxy tasks. Based on the hypothesis: \textit{a weaker correlation means a higher complementarity between two features}, we first calculate the correlation measure (\emph{i.e.,} linear centered kernel alignment (LCKA)~\cite{Kornblith2019CKA}) between the features yielded by different proxy tasks, and then employ the proposed greedy algorithm to iteratively add a proxy task with the lowest LCKA to the proxy tasks in the current task pool, and finally form a multi-task SSL framework. 
Subsequently, we implement a self-aggregation method to enlarge the feature space explored by a proxy task in a mode of \textit{self-complementarity}. To achieve this, an auxiliary loss function based on LCKA is proposed as an add-on to the existing loss function to promote the exploring of where a single proxy task fails to cover.

In summary, our paper contributes two SSL aggregation strategies to exploit the complementarity.
Specifically, 1) our multi-task aggregative SSL (MT-ASSL) provides new insights into multiple proxy task integration and systematically designs a greedy algorithm to combine complementary tasks; and 2) the self-aggregative SSL (Self-ASSL) provides a powerful tool by an add-on complementary feature space for improving the performance of a SSL approach. 
Such aggregative SSL methods boost the robustness of the learned feature representation as demonstrated by our extensive evaluations on a 2D natural image dataset and a 3D medical volume dataset.

\section{Related Works}
Self-supervised learning (SSL), as a new paradigm of unsupervised learning, attracts increasing attentions from the community. The pipeline consists of two steps: 1) pre-train a convolutional neural network (CNN) on a pretext task with an unannotated dataset; 2) fine-tune the pre-trained network for the specific target task with a small set of annotated data (Refer to Section~\ref{math} for mathematical formulation). Recent studies have demonstrated the effectiveness of SSL in computer vision tasks, such as image classification~\cite{gidaris2018image_rotations}, semantic segmentation~\cite{ZHURubik} and object detection~\cite{Zhou2020LookIntoObjectSS, Xie2021DetCoUC, He2020MoCo}. 

As the core of SSL, lots of proxy tasks have been proposed, which can be roughly grouped to image restoration based and ad-hoc prediction based. The former one defines a set of image transformations to permute the original images and requires CNN to restore, \emph{e.g.,} image inpainting~\cite{Pathak2016Inpainting}, colorization~\cite{larsson_colorization_2017, Zhang2016ColorfulIC, Zhang2017SplitBrainAU} and Jigsaw puzzle based approaches~\cite{noroozi2016jigsaw_puzzles, Arbitrary_Jigsaw_Puzzles,Arbitrary_Jigsaw_Puzzles, Mundhenk2018ImprovementsTC}.
The latter one often constructs the self-supervision signal via transformations without contextual permutation. The representative approaches are rotation prediction~\cite{gidaris2018image_rotations}, synthetic/natural image prediction ~\cite{Jenni2018SpotArtifacts} and instance discrimination based on contrastive prediction~\cite{Tian2019ContrastiveMC, chen2020SimCLR, He2020MoCo, Grill2020BYOL, Wu2018UnsupervisedFL, Zhuang2019LocalAF, Caron2020UnsupervisedLO}.
Besides, clustering is also an important direction for unsupervised representation learning \cite{Caron2018DeepClustering, Husser2018AssociativeDC, Chen2019UnsupervisedCO, Ji2019IIC}, which is highly related to SSL. 

As aforementioned, the feature representation learned via a single proxy task leads to a limited generalization on the target tasks.
Hence, researches began to exploit the benefits of integrating multiple proxy tasks via multi-task learning~\cite{Jenni2020LCI, ZHURubik, zhou2019MG}.
Multi-task learning is a prevalent tool using in many computer vision tasks~\cite{Zhang2017SurveyMulti, Kendall2018MultitaskLU}, but few researches investigate its advantages to SSL.
Doersch \emph{et al.}~\cite{Doersch2017MultitaskSSL} firstly attempt to integrate multiple SSL approaches for a better feature representation. Feng \emph{et al.}~\cite{Feng2019SelfSupervisedRL} combine rotation prediction with instances discrimination. However, these proxy task integration approaches are lack of principle investigation. To the best of our knowledge, this is the first study that investigates the aggregation of SSL. 

\begin{figure*}[!th]
\begin{center}
\includegraphics[width=0.9\textwidth]{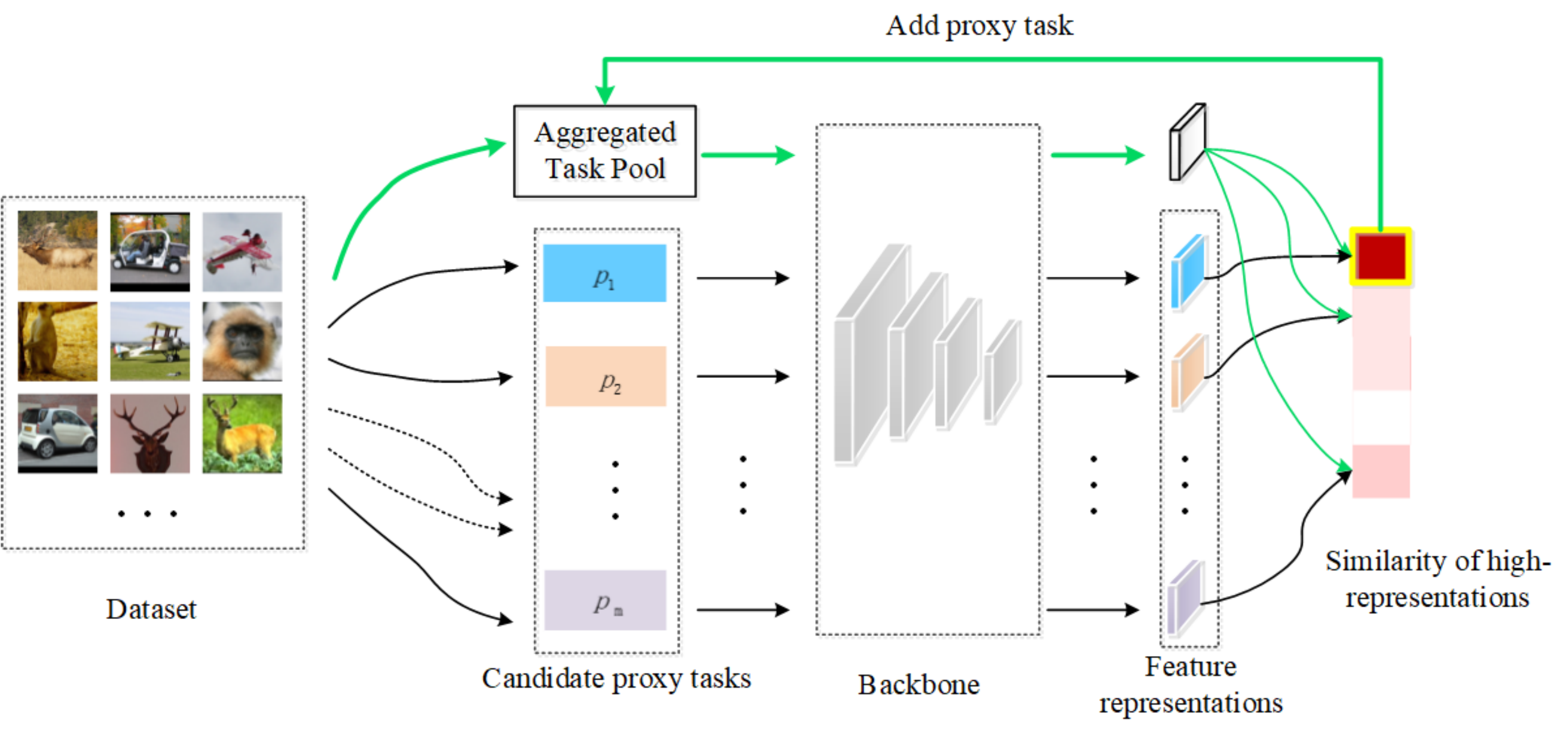}
\end{center}
\caption{\textbf{The proposed multi-task aggregative self-supervised learning strategy (MT-ASSL) of each iteration.} We evaluate the similarity between feature representations learned by aggregated task pool and candidate proxy tasks using LCKA, and accordingly integrate the proxy task with the low similarity into the proposed MT-ASSL framework. The green and black arrows indicate aggregation SSL and single proxy task training.}
\label{figure_MSSL}
\end{figure*}

\section{Formulation of SSL on Limited Samples}\label{math}
We first mathematically illustrate the procedure of SSL training. 
Assume $x=\{x^i\}$ and $y_p=\{y_p^i\}$ are a small group of data and pseudo label of proxy task $p$, respectively, and $x_t=\{x_t^j\}$ and $y_t=\{y_t^j\}$ are the data and labels of target task, respectively. The optimization problem for proxy task $p$ can be written as: 
\begin{equation}\label{Equ:F_p}
\underset{\phi, h_p}  {\min}
F_p(\phi, h_p| x,y_{p}) = \sum_{i}{L_{p}(h_p(\phi(x^i)), y^i_{p})},
\end{equation}
where $\phi$, $h_p$, and $L_p$ denote the backbone model, projection layer and loss function for proxy task, respectively. Following that, $\phi$ is fine-tuned on the target task:
\begin{equation}\label{Equ:F_t}
\underset{\phi, h_t}{\min} F_{t}(\phi, h_t|x_t, y_{t}) =
\sum_{j}
{L_{t}(h_t(\phi(x_t^j)), y^j_{t})},
\end{equation}
where $y_t$, $h_t$, and $L_t$ denote the label, projection layer, and loss function for target task. After the training on target task, the model performance, such as classification accuracy, can be evaluated:
\begin{equation}\label{Equ:ACC}
ACC(\phi|x_t, y_{t})=\sum_{j}{Eval(h_t(\phi(x_t^j)), y^j_{t})},
\end{equation}
where $Eval(.)$ is the specific function for model performance evaluation. The goal of SSL is to obtain a robust $\phi$, which achieves the better performance $ACC$ than the train-from-scratch model via fully exploiting useful information from the small training set $x$.

\section{Multi-Task Aggregative SSL}
Existing studies have demonstrated the effectiveness of multi-task learning for the extraction of a robust feature representation. In this regard, to further boost the robustness of self-supervised feature representations, we propose a principled framework for the integration of multiple SSL approaches.
Assume there is a pool of candidate proxy tasks $\mathcal{P}=\{p_1, p_2, ..., p_{|\mathcal{P}|}\}$, where $|\mathcal{P}|$ denotes the the number of candidate proxy tasks, the conventional multi-task integrates all $\mathcal{P}$ can be formulated as:
\begin{equation}
\begin{aligned}
\underset{\phi}{\min} F_{\mathcal{P}}(\phi|x)=&
\sum_{p\in \mathcal{P}} F_p(\phi, h_p| x,y_{p})
\\=&
\sum_{p\in \mathcal{P}}
\sum_i{L_{p}(h_{p}(\phi(x^i)), y^i_{p})}.
\end{aligned}
\end{equation}

However, since the feature subspaces explored by different proxy tasks are heterogeneous; they might highly overlap or isolated. A blind integration of such feature subspaces brings limited (if not damaged) performance improvement. 
To address the problem, we systematically explore the feature complementarity between multiple SSL approaches, and propose a novel multi-task aggregative self-supervised learning (MT-ASSL) strategy to \textbf{iteratively integrate} the features extracted by different-yet-complementary proxy tasks. 
The objective of our MT-ASSL can be defined as:
\begin{equation}\label{Equ:F_{MT}}
\underset{\phi, \mathcal{A}}{\min}
F_{MT}(\phi|x)=
\sum_{p\in \mathcal{A}} F_p(\phi, h_p| x,y_{p})
\end{equation}
where $\mathcal{A}$ denotes the selected subset of proxy tasks for aggregation, \textit{which is optimized} too. 

In initial stage, for each $p\in \mathcal{P}$ we train the network from scratch according to Eq.~(\ref{Equ:F_p}) to obtain its representation  $\phi_p$, which is then fined-tuned based on Eq.~(\ref{Equ:F_t}). Suppose that task ${\hat p}$ has with the highest evaluating performance of target task,
\begin{equation}\label{Equ:Ainit}
{\hat p} =
\underset{p\in \mathcal{P}}{argmax}
ACC(\phi_p|x_t, y_t),
\end{equation}
then it is added to $\mathcal{A}=\{{\hat p}\}$, and removed from $\mathcal{P}$. 
The initial aggregation feature representation is set as  $\phi_{\mathcal{A}}  = \phi_{\hat p}$.

We then attempt to add into $\mathcal{A}$ with more proxy tasks from $\mathcal{P}$ in order to learn a stronger aggregation feature representation. We follow a greedy algorithm by adding one proxy task at one iteration.  
To achieve so, we leverage the hypothesis underlying our MT-ASSL, that is, \textit{a weaker correlation means a higher complementarity among features}. 

To measure the complementarity between features learned by two different proxy task, we choose to use 
\textbf{linear centered kernel alignment (LCKA)} proposed by Kornblith et al.~\cite{Kornblith2019CKA}, who conducted a careful study on measuring similarity between the representations learned by neural networks, among other choices~\cite{Wilks2011CCA, Morcos2018InsightsOR}. Mathematically, the LCKA between two feature representations, say $\phi_\mathcal{A}$ and $\phi_p$, are defined as 
\begin{equation}\label{Equ:S_p}
    S[\phi_\mathcal{A}, \phi_p]=\frac{\|{G[\phi_\mathcal{A}] \odot G[\phi_p]}\|_1}{\sqrt{\|{ G[\phi_\mathcal{A}] \|_1 \|G[\phi_p]\|_1}}},
\end{equation}
\begin{equation}\label{Equ:S_p2}
    G[\phi] = C(\phi(x)^T \phi(x)), 
\end{equation}
where $\phi(x)$ is a matrix that contains the features for all samples in $x$, $(.)^T$ and $\odot$ denote matrix multiplication and element-wise multiplication, respectively; $C(.)$ denotes the centered alignment operation, which is defined as:
\begin{equation}
C(M)=M(I_d-{\textbf{11}}^T/d),  
\end{equation} 
where $d$ denotes the dimension of a square matrix $M_{d\times d}$, $I_d$ is an identity matrix of size $d \times d$, and $\textbf{1}$ is a $d \times 1$ vector of ones.

\begin{algorithm}[t]  
\caption{Multi-task aggregative SSL}
\label{alg:Framwork}  
\begin{algorithmic}[1] 
\STATE {\bf Input:}
\STATE \quad A small group of data $x$ of proxy task;
\STATE \quad A small group of data $x_t$ and label $y_t$ of target task;
\STATE \quad Backbone model $\phi$;
\STATE \quad Candidate proxy task list: $\mathcal{P}=\{p_1, p_2, ..., p_{|\mathcal{P}|}\}$;
\STATE {\bf Procedure:}
\STATE Aggregated$\_$Task$\_$Pool: $\mathcal{A}$ = []
\FOR{$p$ in $\mathcal{P}$:} 
    \STATE Optimize $F_p(\phi_p,h_p|x,y_p)$ by Eq.~(\ref{Equ:F_p}) to obtain $\phi_p$
    \STATE Optimize $F_t(\phi_p, h_t,|x_t,y_t)$ by Eq.~(\ref{Equ:F_t}) to obtained an updated  $\phi_p$ 
    \STATE Obtain $ACC(\phi_p| x_t, y_t)$ by Eq.~(\ref{Equ:ACC})
\ENDFOR
\STATE Obtain ${\hat p}$ by Eq.~(\ref{Equ:Ainit})
\STATE $\mathcal{A}\leftarrow \{{\hat p}\}$
\STATE $P\leftarrow P- \{{\hat p}\}$
\STATE Best$\_$Acc = $ACC(\phi_{\hat p}|x_t, y_t)$ 
\WHILE {Best$\_$Acc is updated and $\mathcal{P}\neq[]$}
    \STATE obtain $S[\phi_\mathcal{A}, \phi_p]$ for each $ p\in \mathcal{P}$ by Eq.~(\ref{Equ:S_p})
    \STATE obtain $p_{x}$ by Eq.~(\ref{Equ:px})
    \STATE $\mathcal{A} \leftarrow \mathcal{A} \cup \{p_x\}$
    \STATE $\mathcal{P} \leftarrow  \mathcal{P}- \{p_x\}$
    \STATE Optimize $F_{MT}(\phi_{\mathcal{A}}|x)$ by Eq.~(\ref{Equ:F_{MT}}) and obtain $\phi_{\mathcal{A}}$
    \STATE Optimize $F_t(\phi_{\mathcal{A}}, h_t,|x_t,y_t)$ by Eq.~(\ref{Equ:F_t}) to obtained an updated $\phi_{\mathcal{A}}$ 
    \STATE Obtain $ACC_{\mathcal{A}}=ACC(\phi_{\mathcal{A}}|x_t, y_t)$ by Eq.~(\ref{Equ:ACC})
    \STATE Best$\_$ACC = $ACC_{\mathcal{A}}$ if $ACC_{\mathcal{A}} \geq $ Best$\_$ACC
\ENDWHILE
\STATE {\bf Output:} Aggregation task pool $\mathcal{A}$
\STATE \quad \quad \quad \quad Trained backbone model $\phi_{\mathcal{A}}$.
\end{algorithmic}  
\end{algorithm}

LCKA offers a simple yet concrete measurement of task correlation, which can be used as the guideline for aggregation of multiple proxy tasks. We update the task pool $\mathcal{A}$ by adding one more task that has the lowest correlation with the existing tasks in the pool:
\begin{equation}\label{Equ:px}
p_x=
\underset{p\in \mathcal{P}}{\arg \min}
S[\phi_\mathcal{A}, \phi_p].
\end{equation}
\begin{equation}\label{Equ:Aiter}
\mathcal{A}=\mathcal{A} \cup \{p_x\}
\end{equation}
The selected $p_x$ is removed from $\mathcal{P}$. 

Following this selection criterion, a greedy training strategy is proposed to aggregate different proxy tasks for feature learning in an iterative fashion (from Eqs.~(\ref{Equ:F_{MT}}) to (\ref{Equ:Aiter})) The algorithm stops as long as there is no performance gain by adding a new proxy task
or $\mathcal{P}$ is blank. Therefore, as the training iteration increases, the features exploited by different proxy tasks are gradually integrated, which yields a feature representation of better generalization. The process of the proposed aggregative training is described in Algorithm~\ref{alg:Framwork}.

\section{Self-Aggregative SSL}
Since our MT-ASSL is in virtue of different SSL methods, we further propose a novel aggregation strategy based on \textbf{self-complementarity}, namely self-aggregative SSL (Self-ASSL), to boost the generalization of the feature, going beyond the limitation of a single SSL. 

\begin{figure}[!htb]
\begin{center}
 \includegraphics[width=0.88\linewidth]{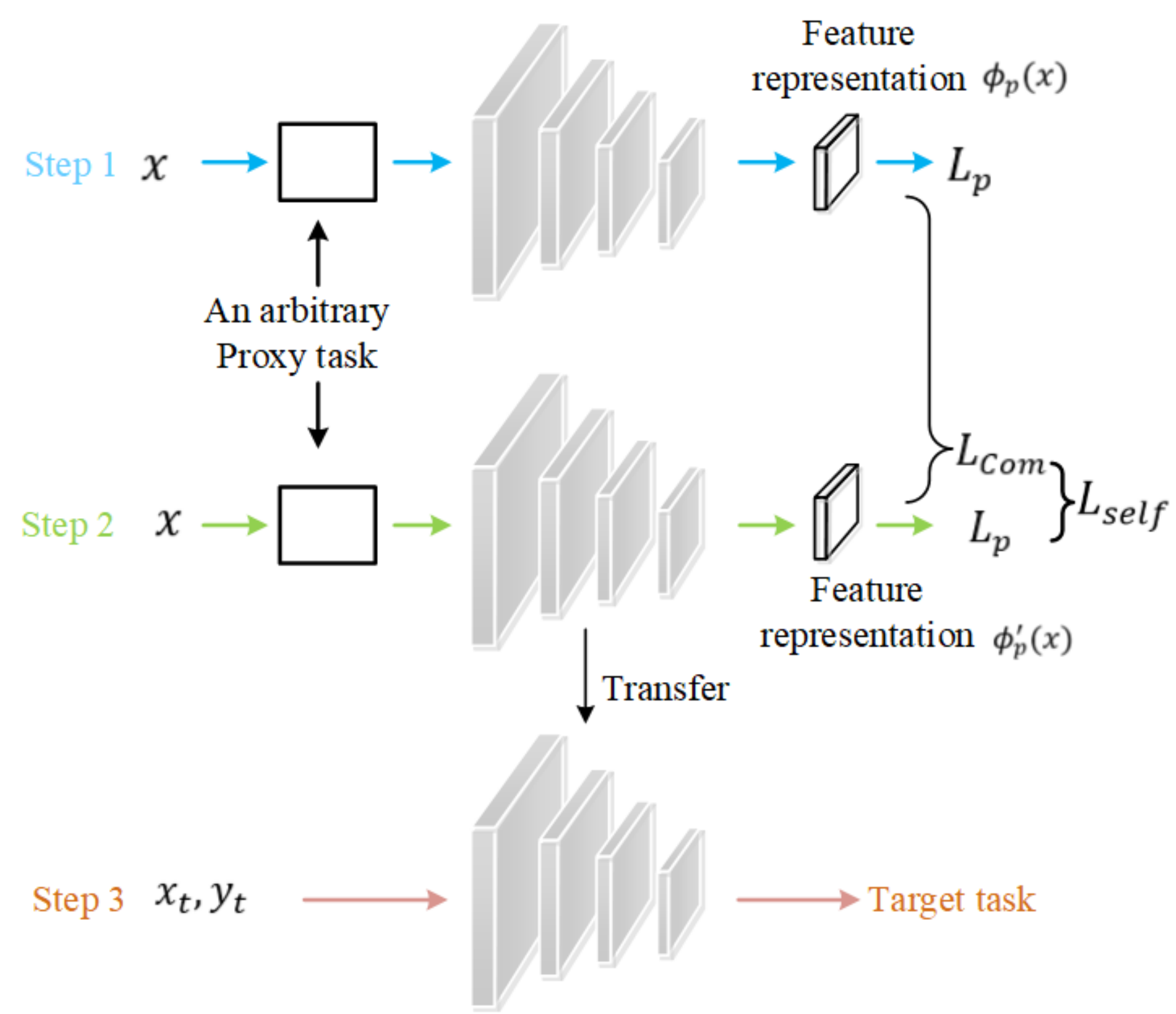}
\end{center}
\caption{\textbf{The proposed self-aggregative SSL (Self-ASSL) learning pipeline.} It includes three different training steps. The feature representation $\phi'_p(x)$ gains a better generalization by aggregating a latent space complement to $\phi_p(x)$.}
\label{figure_ComSSL}
\end{figure}

The pipeline of our Self-ASSL is shown in Figure~\ref{figure_ComSSL}, which includes three steps. First, we train the backbone network by Eq.~(\ref{Equ:F_p}) to obtain a feature representation $\phi_p(x)$. Then in Step 2 of Figure~\ref{figure_ComSSL}, we attempt to learn a new, self-complementary representation $\phi'_{p}(x)$, which ideally should have a low similarity with $\phi_p(x)$, using the following objective:
\begin{equation}\label{Equ:F_self}
\begin{aligned}
\underset{\phi’_p, h'_p}{\min} F_{Self}(\phi’_p, h'_p|\phi_p,x, y_{p}) = L_{self} + L_{com};
\end{aligned}
\end{equation}
\begin{equation}
L_{self}=\sum_{i=1}{L_{p}(h'_p(\phi'_p(x^i)), y^i_{p})}, L_{com} =
-S[\phi'_p,\phi_p ].
\end{equation}
where the trained backbone $\phi_p$ is frozen. After several iterations, the model $\phi'_{p}$, achieving the better generalization, can further boost the improvement of the subsequent target task, by optimizing $F_t(\phi'_p,h_t|x_t,y_t)$ in Eq.~(\ref{Equ:F_t}).

\begin{table*}[!th]
\centering
\caption{\textbf{Evaluation of the effectiveness of proxy task integration that aggregates each pair of SSLs (`A1' and `A2') on STL10 dataset.} Proxy task `A2' is sorted by the similarity of the last layer of ResNet-18 to `A1'. The `Avg ACC' and `Max ACC' are the average ACC and the maximum ACC calculated by separately fine-tuning 'A1' and 'A2' pre-trained weights on the target task. The accuracy of proxy task integration is denoted as `Int ACC'. Avg (+/-) and Max (+/-) are the improvements comparing Int ACC with Avg ACC and Max ACC, respectively. The numbers in blue and red represent the top 1 and 2 records among the whole column, respectively. Similarly, the italic numbers represent the top 1 improvements for the SSL combinations.}
\begin{tabular}{ccc|cc|ccc}
\hline
A1        & A2        & Similarity & Avg ACC & Max ACC & Int ACC & Avg (+/-) & Max (+/-) \\\hline\hline
\multirow{5}{*}{2D Jigsaw} & 2D Rot    & 0.1543     & 69.18   & 70.07   & 76.06    & \underline{+6.88}      & \underline{+5.99}      \\
 & SRC       & 0.1891     & 68.33   & 68.36   & 75.57    & \textit{\textbf{+7.25}}      & \textit{\textbf{+7.21}}      \\
 & SimCLR    & 0.2409     & \textcolor{blue}{\textit{70.67}}   & \textit{\textbf{73.05}}   & 74.69    & +4.02      & +1.64      \\
 & Inpaint   & 0.4132     & 66.39   & 68.29   & 70.77    & +4.38      & +2.48      \\
 & MCPC      & 0.4202     & 67.62   & 68.29   & 68.81    & +1.20      & +0.52      \\\hline
\multirow{5}{*}{2D Rot}    & SRC       & \textit{\textbf{0.0695}}     & 69.22   & 70.07   & \textcolor{blue}{76.70}    & \textit{\textbf{+7.49}}      & \textit{\textbf{+6.63}}      \\
    & SimCLR    & \textcolor{red}{0.0866}     & \textit{\textbf{71.56}}   & \textit{\textbf{73.05}}   & \textit{\textbf{78.21}}    & +6.65      & +5.16      \\
    & Inpaint   & 0.1524     & 67.28   & 70.07   & 73.44    & +6.16      & +3.37      \\
    & 2D Jigsaw & 0.1543     & 69.18   & 70.07   & 76.06    & \underline{+6.88}      & \underline{+5.99}      \\
    & MCPC      & 0.1557     & 68.51   & 70.07   & 74.70    & +6.20      & +4.63      \\\hline
\multirow{5}{*}{SimCLR}    & 2D Rot    & \textcolor{red}{0.0866}     & \textit{\textbf{71.56}}   & \textit{\textbf{73.05}}   & \textit{\textbf{78.21}}    & \textit{\textbf{+6.65}}      & \textit{\textbf{+5.16}}      \\
    & SRC       & \textcolor{blue}{0.1085}     & \textcolor{red}{70.71}   & \textit{\textbf{73.05}}   & \textcolor{red}{77.25}    & \underline{+6.55}      & \underline{+4.20}      \\
    & Inpaint   & 0.2373     & 68.77   & \textit{\textbf{73.05}}   & 74.08    & +5.31      & +1.03      \\
    & MCPC      & 0.2388     & 70.00   & \textit{\textbf{73.05}}   & 73.29    & +3.30      & +0.24      \\
    & 2D Jigsaw & 0.2409     & \textcolor{blue}{70.67}   & \textit{\textbf{73.05}}   & 74.69    & +4.02      & +1.64      \\\hline
\multirow{5}{*}{Inpaint}   & 2D Rot    & 0.1524     & 67.28   & 70.07   & 73.44    & \textit{\textbf{+6.16}}      & \underline{+3.37}      \\
   & SRC       & 0.1852     & 66.43   & 68.36   & 72.47    & \underline{+6.04}      & \textit{\textbf{+4.11}}      \\
   & SimCLR    & 0.2373     & 68.77   & \textit{\textbf{73.05}}   & 74.08    & +5.31      & +1.03      \\
   & MCPC      & 0.4125     & 65.72   & 66.94   & 65.09    & -0.63     & -1.85     \\
   & 2D Jigsaw & 0.4132     & 66.39   & 68.29   & 70.77    & +4.38      & +2.48      \\\hline
\multirow{5}{*}{SRC}       & 2D Rot    & \textit{\textbf{0.0695}}     & 69.22   & 70.07   & \textcolor{blue}{76.70}    & \textit{\textbf{+7.49}}      & \underline{+6.63}      \\
       & SimCLR    & \textcolor{blue}{0.1085}     & \textcolor{red}{70.71}   & \textit{\textbf{73.05}}   & \textcolor{red}{77.25}    & +6.55      & +4.20      \\
       & Inpaint   & 0.1852     & 66.43   & 68.36   & 72.47    & +6.04      & +4.11      \\
       & 2D Jigsaw & 0.1891     & 68.33   & 68.36   & 75.05    & \underline{+6.73}      & \textit{\textbf{+6.69}}      \\
       & MCPC      & 0.2573     & 67.65   & 68.36   & 72.03    & +4.38      & +3.67      \\\hline
\multirow{5}{*}{MCPC}      & 2D Rot    & 0.1557     & 68.51   & 70.07   & 74.70    & \textit{\textbf{+6.20}}      & \textit{\textbf{+4.63}}      \\
      & SimCLR    & 0.2388     & 70.00   & \textit{\textbf{73.05}}   & 73.29    & +3.30      & +0.24      \\
      & SRC       & 0.2573     & 67.65   & 68.36   & 72.03    & \underline{+4.38}      & \underline{+3.67}      \\
      & Inpaint   & 0.4125     & 65.72   & 66.94   & 65.09    & -0.63     & -1.85     \\
      & 2D Jigsaw & 0.4202     & 67.62   & 68.29   & 68.81    & +1.20      & +0.52    \\\hline\hline
\end{tabular}
\label{table_ASSL_STL10_2}
\end{table*}

\begin{table*}[!th]
\centering \small
\caption{\textbf{Results of our multi-task aggregative SSL (MT-ASSL) on STL10 dataset.} The `A1' and `A2' indicate two single SSL methods for aggregation. `A2' are sorted by the similarity of the fourth layer to proxy task `A1'. The `Avg ACC' and `Max ACC' are the average ACC and the maximum ACC of `A1' and `A2'. Avg (+/-) and Max (+/-) are the improvements comparing MT-ASSL ACC with Avg ACC and Max ACC, respectively. The bold and italic numbers represent the top 1 and 2 records among the sub-column, respectively. (Iter.--Iteration)}
\begin{tabular}{c|ccc|cc|ccc}
\hline
Iter. & A1        & A2        & Similarity & Avg ACC & Max ACC & MT-ASSL ACC & Avg (+/-) & Max (+/-) \\\hline\hline
\multirow{6}{*}{1}    & SRC       & -  &  &  &  & 68.36 &  &  \\
    & 2D Jigsaw & -  &  &  &  & 68.29 &  &  \\
    & Inpaint   & -  &  &  &  & 64.49 &  &  \\
    & SimCLR    & -  &  &  &  & \textit{\textbf{73.05}} &  &  \\
    & 2D Rot    & -  &  &  &  & \underline{70.07} &  &  \\
    & MCPC      & -  &  &  &  & 66.94 &  &  \\\hline\hline
\multirow{5}{*}{2}    & \multirow{5}{*}{SimCLR}    & 2D Rot    & \textit{\textbf{0.0866}}     & \textit{\textbf{71.56}}   & \textit{\textbf{73.05}}   & \textit{\textbf{78.21}}    & \textit{\textbf{+6.65}}      & \textit{\textbf{+5.16}}      \\
    &    & SRC       & \underline{0.1085}     & \underline{70.71}   & \textit{\textbf{73.05}}   & \underline{77.25}    & \underline{+6.55}      & \underline{+4.20}      \\
    &    & Inpaint   & 0.2373     & 68.77   & \textit{\textbf{73.05}}   & 74.08    & +5.31      & +1.03      \\
    &    & MCPC      & 0.2388     & 70.00   & \textit{\textbf{73.05}}   & 73.29    & +3.30      & +0.24      \\
    &    & 2D Jigsaw & 0.2409     & 70.67   & \textit{\textbf{73.05}}   & 74.69    & +4.02      & +1.64      \\\hline\hline

\multirow{4}{*}{3}    & \multirow{4}{*}{SimCLR + 2D Rot} & SRC       & \textit{\textbf{0.0911}} & 73.29 & 78.21 & \textit{\textbf{79.43}} & \underline{+6.15} & \textit{\textbf{+1.22}}  \\
    &                           & Inpaint   & \underline{0.1973} & 71.35 & 78.21 & \underline{78.16} & \textit{\textbf{+6.81}} & -0.05 \\
    &                         &  MCPC      & 0.1985 & 72.58 & 78.21 & 77.97 & +5.40 & -0.24 \\
    &                                & 2D Jigsaw & 0.1986 & 73.25 & 78.21 & 77.89 & +4.64 & -0.32      \\\hline\hline
\multirow{3}{*}{4}  & \multirow{3}{*}{SimCLR + 2D Rot + SRC} & Inpaint   & \textit{\textbf{0.2251}} & 71.96 & \textit{\textbf{79.43}} & 76.01 & +4.05 & -3.42 \\
    & & MCPC      & \underline{0.2283} & 73.19 & \textit{\textbf{79.43}} & 75.78 & +2.60 & -3.65 \\
    & & 2D Jigsaw & 0.2284 & 73.86 & \textit{\textbf{79.43}} & 76.25 & +2.39 & -3.18 \\\hline\hline

\end{tabular}
\label{table_ASSL_STL10}
\end{table*}

\begin{table*}[!th]
\centering
\caption{\textbf{Results of our multi-task aggregative SSL (MT-ASSL) on brain hemorrhage dataset.} Mark explanation is the same as Table~\ref{table_ASSL_STL10}.}
\label{table_ASSL_brain}
\begin{tabular}{c|ccc|cc|ccc}
\hline
Iter. & A1     & A2     & Similarity & Avg ACC & Max ACC & MT-ASSL ACC & Avg (+/-) & MAX (+/-) \\\hline\hline
\multirow{5}{*}{1}    & SC-ASSL     & - &  &  &  & \textit{\textbf{89.53}} &  &  \\
& Cube   & - &  &  &  & \underline{87.50} &  &  \\
& 3D CPC & - &  &  &  & 83.79 &  &  \\
& 3D Rot & - &  &  &  & 87.16 &  &  \\
& MG     & - &  &  &  & 87.50 &  & \\\hline\hline
\multirow{4}{*}{2}    & \multirow{4}{*}{SC-ASSL}     & Cube   & \textit{\textbf{0.0612}}     & \textit{\textbf{88.52}}   & \textit{\textbf{89.53}}   & \textit{\textbf{90.20}}    & \textbf{+1.69}      & \textbf{+0.67}      \\
     & & 3D CPC & \underline{0.1722}     & 86.66   & \textit{\textbf{89.53}}   & 87.50    & \underline{+0.84}      & -2.03     \\
     & & 3D Rot & 0.2413     & \underline{88.35}   & \textit{\textbf{89.53}}   & 88.17    & -0.17     & -1.36     \\
     & & MG     & 0.2415     & \textit{\textbf{88.52}}   & \textit{\textbf{89.53}}   & 87.83    & -0.69     & -1.70     \\\hline\hline
\multirow{3}{*}{3}    &\multirow{3}{*}{SC-ASSL + Cube} & 3D Rot & \textit{\textbf{0.0403}}     & 88.68   & \textit{\textbf{90.20}}   & 89.52    & +0.84      & -0.68     \\
                          &  & MG     & \underline{0.1054}     & 88.85   & \textit{\textbf{90.20}}   & 88.51    & -0.34     & -1.69     \\
                          &  & 3D CPC & 0.1059     & 87.00   & \textit{\textbf{90.20}}   & 87.50    & +0.50      & -2.70    \\\hline
\end{tabular}
\end{table*}

\begin{table*}[!t]
\centering \footnotesize
\caption{\textbf{Accuracy (ACC \%) of different proxy tasks tested on STL10 and Brain hemorrhage datasets.} ACC (+/-) lists the improvements of ACC comparing Self-ASSL-trained SSLs to the original ones. The bold and italic numbers represent the top 1 and 2 records among the sub-column, respectively. (T. f. s.--Train from scratch)}
\begin{tabular}{llccc|llccc}
\hline
\multicolumn{5}{c|}{STL10}                                        & \multicolumn{5}{c}{Brain hemorrhage}                              \\\hline
Backbone & Method  & ACC   & w/ Self-ASSL & ACC (+/-) & 
Backbone & Method  & ACC   & w/ Self-ASSL & ACC (+/-) \\\hline\hline
\multirow{8}{*}{2D VGG}      & T. f. s.   & 63.90 & -        & -     & \multirow{8}{*}{3D VGG}      & T. f. s. & 72.30 & -        & -     \\
                             & 2D Jigsaw & 62.84 & 63.82    & +0.98  &                              & 3D CPC  & 77.02 & 83.44    & \textbf{+6.42}  \\
                             & Inpaint   & 64.76 & 65.25    & +0.49  &                              & 3D Rot  & 76.68 & 79.05    & +2.37  \\
                             & 2D Rot    &\underline{69.33} & \underline{69.09}  &  -0.24     &                              & Cube    & 77.36 & 81.08    & \underline{+3.72}  \\
                             & SimCLR    & \textbf{70.31} & \textbf{73.04}  & \underline{+2.73}  &                              & MG      & 85.81 & 86.15    & +0.34  \\
                             & SRC         & 62.90 & 63.21 & +0.31 &  & SRC  & 85.81 & 87.50 & +1.69 \\
                             & MCPC & 65.25 & 68.42 & \textbf{+3.17} &  & MCPC & \underline{87.50} & \textbf{89.52} & +2.02 \\
                             & 2D SC-ASSL    & 65.69 & 66.45    & +0.76  &                              & 3D SC-ASSL  & \textbf{87.83} & \underline{88.17}    & +0.34  \\\hline
\multirow{8}{*}{2D ResNet-18} & T. f. s.   & 63.19 & -        & -     & \multirow{8}{*}{3D ResNet-18} & T. f. s. & 81.08 & -        & -     \\
                             & 2D Jigsaw & 68.29 & 70.80    & \textbf{+2.51}  &                              & 3D CPC  & 83.79 & 88.17    & \textbf{+4.38}  \\
                             & Inpaint   & 64.49 & 65.67    & +1.18  &                              & 3D Rot  & 85.81 & 85.47    & -0.34     \\
                             & 2D Rot    & 70.07 & 72.41    & \underline{+2.34}  &                              & Cube    & 87.50 & \underline{88.85}    & \underline{+1.35}  \\
                             & SimCLR    & \textbf{73.05} & \underline{73.38}    & +0.33  &                              & MG      & 87.50 & 88.17    & +0.67  \\
                              & SRC  & 68.36 & 69.33 & +0.97 & & SRC  & 87.16 & 88.51 & \underline{+1.35} \\
                              & MCPC & 66.94 & 67.18 & +0.24 & & MCPC & \underline{88.51} & \underline{88.85} & +0.34 \\
                             & 2D SC-ASSL    & \underline{72.03} & \textbf{74.28}    & +2.25  &                              & 3D SC-ASSL  & \textbf{89.53} & \textbf{89.53}    & +0.00 \\\hline
\end{tabular}
\label{table:CSSL}
\vspace{-3mm}
\end{table*}

\section{Experiments}
In this section, we conduct extensive experiments on datasets with a limited number of samples to validate the effectiveness of the proposed self-supervised learning paradigms (i.e. MT-ASSL and Self-ASSL).

\paragraph{2D natural image dataset: STL10.} The STL10 dataset is collected for image classification. Concretely, the dataset has a handful of annotation (10 classes, 500 per class) and provides a large amount of unlabeled images (100K) for the development of unsupervised learning approaches. The SSL methods are first trained on unlabeled data and then fine-tuned on 5K labeled data for classification task. All the images are of a uniform size $96\times 96$ pixels

\paragraph{3D medical volume dataset: Brain hemorrhage.} The brain hemorrhage dataset, containing 1,486 brain CT volumes, is collected from our collaborative hospital with IRB approval. The CT volumes can be separated to four classes according to the pathological causes of cerebral hemorrhage: aneurysm, arteriovenous malformation, moyamoya disease and hypertension. The size of CT volumes is standardized to $30 \times 270 \times 230$ voxels. We separate the brain hemorrhage dataset into training and testing sets according to the ratio of 80:20. 

\paragraph{Implementation details.}
For 2D natural image classification, several state-of-the-art SSL approaches, including SRC~\cite{anonymous}, MCPC~\cite{anonymous}, 2D jigsaw puzzles (2D Jigsaw)~\cite{noroozi2016jigsaw_puzzles}, image rotation (2D Rot)~\cite{gidaris2018image_rotations}, image inpainting (Inpaint)~\cite{Pathak2016Inpainting} and SimCLR~\cite{chen2020SimCLR}, are involved to form a pool of proxy tasks for our MT-ASSL and Self-ASSL. The 2D ResNet-18~\cite{He2016ResNet} is adopted as backbone network for MT-ASSL and Self-ASSL.
For 3D SSL method, several 3D-based SSL approaches are included, \emph{i.e.,} 3D SC-ASSL~\cite{anonymous}, 3D rotation (3D Rot)~\cite{gidaris2018image_rotations}, Model genesis (MG)~\cite{zhou2019MG}, 3D CPC (3D version of ~\cite{henaff2019dataefficient}) and Rubik's cube (Cube)~\cite{Zhuang_2019_MICCAI}. The 3D ResNet-18 is utilized as backbone for MT-ASSL and Self-ASSL.

All the SSL methods are implemented using PyTorch. We only use random horizontal flip for data augmentation. The network is trained with a mini-batch size of 256 and 16 for STL10 and brain hemorrhage datasets, respectively. The initial learning rate for the proxy task and target task are set to $1e^{-3}$ and $1e^{-4}$ for STL10, $2e^{-5}$ and $15e^{-6}$ for brain hemorrhage dataset. The Adam solver \cite{kingma2014adam} is used as the optimizer for network training. The average classification accuracy (ACC) is employed as the metric for evaluation.

\subsection{Evaluation of MT-ASSL}
To demonstrate the effectiveness of integrating different proxy tasks and explore the performance improvement caused by embedding the complementary information, we conduct a simple experiment on the STL10 dataset --- evaluating the integration performance of two proxy tasks (denoted as $A1$ and $A2$) randomly selected from the SSL pool. The evaluation results are presented in Table~\ref{table_ASSL_STL10_2}. We first pre-train and fine-tune the ResNet-18 using each of the two proxy tasks and record the accuracy, i.e., the average accuracy (Avg ACC) of the paired proxy tasks and the maximum accuracy (Max ACC) among them.
Then, we pre-train and fine-tune another ResNet-18 simultaneously using the two proxy tasks. The aggregation is achieved by iteratively optimizing the loss functions of $\mathcal{L}_{A1}$ and $\mathcal{L}_{A2}$.

It can be easily observed from Table~\ref{table_ASSL_STL10_2} that the feature learned by proxy task integration (Int ACC) yields consistent improvements to the classification accuracy, compared to the single one. Furthermore, we evaluate the similarity between the feature representations learned by different proxy tasks using LCKA, as presented in the `Similarity' column of Table~\ref{table_ASSL_STL10_2}. The performance improvements are observed to decline as the similarity increases, which confirms our hypothesis: \textit{a weaker correlation or similarity means a higher complementarity between two features}.
Hence, the feature learned by the aggregation of proxy tasks with low similarity (high complementarity) can significantly boost the target classification accuracy. Here we further validate the effectiveness of the proposed MT-ASSL strategy that integrates multiple proxy tasks using both 2D and 3D datasets.

\vspace{-10pt}
\paragraph{STL10.} We first evaluate the proposed MT-ASSL on STL10 dataset. 
The results are shown in Table~\ref{table_ASSL_STL10}, following the greedy algorithm as presented in Algorithm~\ref{alg:Framwork}. 
For the first iteration, SimCLR, which achieves the best transferring performance of $73.05\%$, is added to the aggregation task pool. The 2D Rot and SRC proxy tasks with a lower similarity to the aggregation task pool are involved during iteration two and three, respectively. The MT-ASSL is completed after four iterations since no further performance improvement is observed as the similarities of the rest three proxy tasks are nearly the same. 
Therefore, our MT-ASSL obtains the best combination of proxy tasks (\emph{i.e.,} SimCLR + 2D Rot + SRC) for the image classification on STL10 dataset, which results in a final target accuracy of $79.43\%$. 

We notice that the experimental results in this study are lower than those reported in ~\cite{Ji2019IIC, Jenni2020LCI}. Such a degradation is caused by the data augmentation method and backbone network adopted in different approaches. The main purpose of our experiments is to demonstrate that the proposed principled multi-task aggregation approach can effectively exploit feature complementarity between different proxy tasks and significantly surpass the single one. However, the diverse data augmentation \cite{Ji2019IIC} and powerful backbone \cite{Jenni2020LCI} may naturalize the feature complementarity. This is the reason why we augment the data only with the simple horizontal flipping operation and adopt the ResNet-18 as the backbone. Furthermore, since these two approaches adopt different paradigms from the conventional self-supervised proxy tasks, \emph{i.e.,} a generative adversarial network (GAN) and an architecture of multiple over-clustering heads are utilized in \cite{Jenni2020LCI} and \cite{Ji2019IIC}, respectively. Hence, these two approaches are excluded from our proxy task pool.



\begin{figure*}[!th]
\begin{center}
\includegraphics[width=1.\linewidth]{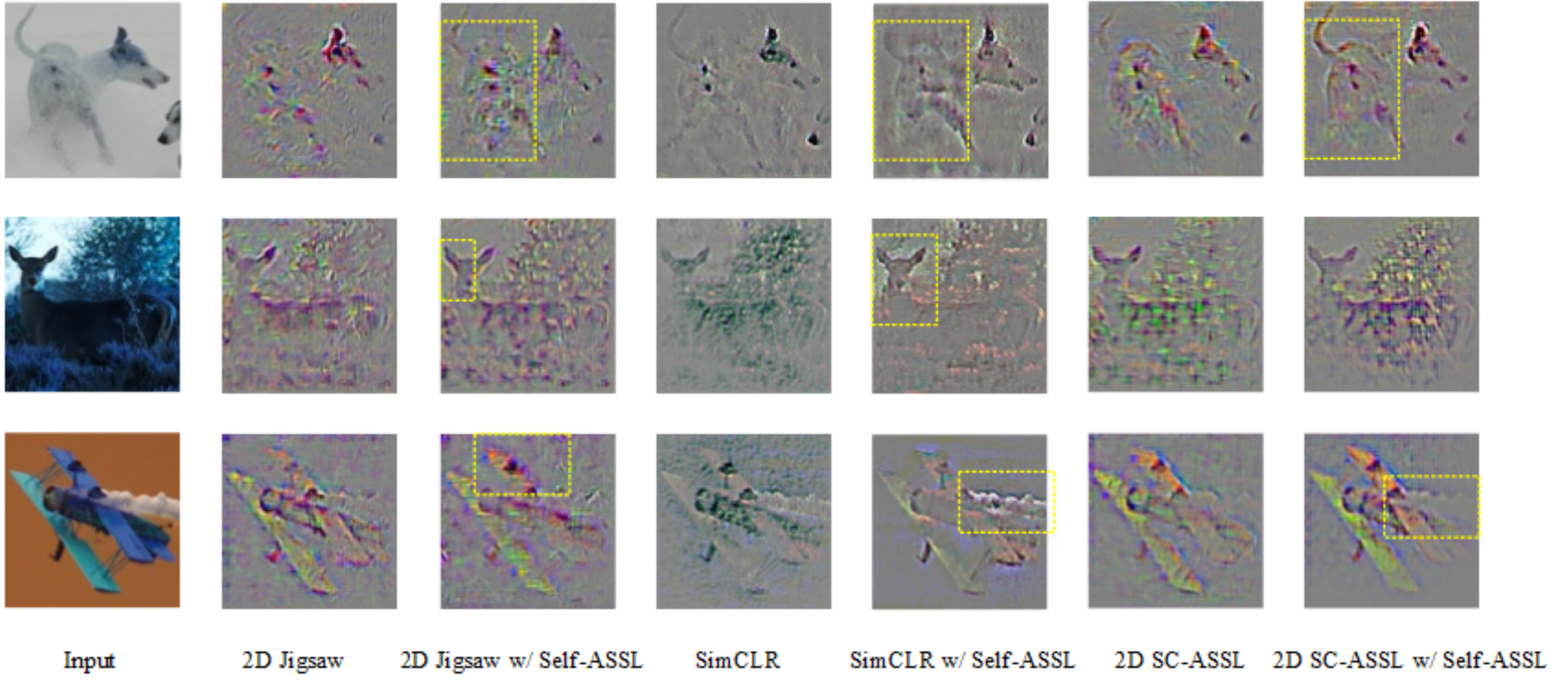}
\end{center}
\caption{\textbf{Grad-CAM visualization of SSL and Self-ASSL.} The yellow rectangles indicate the pronounced differences. 
}
\label{figure_CAM}
\vspace{-2mm}
\end{figure*}

\paragraph{Brain hemorrhage.} We also evaluate our MT-ASSL with five 3D-based SSL methods on brain hemorrhage dataset. The evaluation results are in Table~\ref{table_ASSL_brain}.
The best model is the combination of SC-ASSL and Cube. The aggregation of 3D CPC is observed to degrade the performance by a large margin of $-2.70\%$. The underlying reason is that SC-ASSL also contains a modified CPC branch; therefore, the information exploited by 3D CPC may be redundant to the integration of SC-ASSL and Cube, which deteriorates the pre-training.

\subsection{Evaluation of Self-ASSL}
For the evaluation of Self-ASSL, apart from the ResNet-18 model, we also use the VGG~\cite{Simonyan2015VGG} model as a backbone to validate the generalization of the proposed approaches. The evaluation on the STL10 and brain hemorrhage datasets are presented in Table~\ref{table:CSSL}.

For STL10, it is observed that the Self-ASSL strategy consistently boosts the accuracy of proxy tasks, \emph{e.g.,} $+2.73\%$ for SimCLR with VGG and $+2.51\%$ for 2D Jigsaw with ResNet-18, with the only exception of 2D Rot with VGG. The exact reason of such an exception is unclear and worthy of further investigation. 
A similar trend of improvement is observed on the brain hemorrhage dataset. Our Self-ASSL training strategy boosts the 3D CPC and Cube with VGG by large margins of $+6.42\%$ and $+3.72\%$, respectively. Also, the only exception happens to 3D Rot with ResNet-18. The SC-ASSL with Self-ASSL outperforms the benchmarking algorithms on brain hemorrhage dataset, \emph{i.e.,} an ACC of $89.53\%$ is achieved using ResNet-18 as backbone.

\vspace{2mm}
\noindent{\bf Visualization.} To further demonstrate the effectiveness of our Self-ASSL, we employ Guided Grad-CAM~\cite{Selvaraju2019GradCAM} to visualize the feature learned by the last convolution layer of ResNet-18. Three examples are presented in Figure~\ref{figure_CAM}. The Self-ASSL brings more attention to detailed information (marked using yellow rectangles), which is ignored by the conventional SSL method. For example, the Jigsaw with Self-ASSL captures the information of the dog body, which is omitted by the original Jigsaw. Overall, the feature visualization further validates the effectiveness of our Self-ASSL in helping the proxy task to capture more detailed information by aggregating self-complementary features from a raw image, thereby leading to improved classification accuracy.

\begin{figure}[hb]
\begin{center}
 \includegraphics[width=1.0\linewidth]{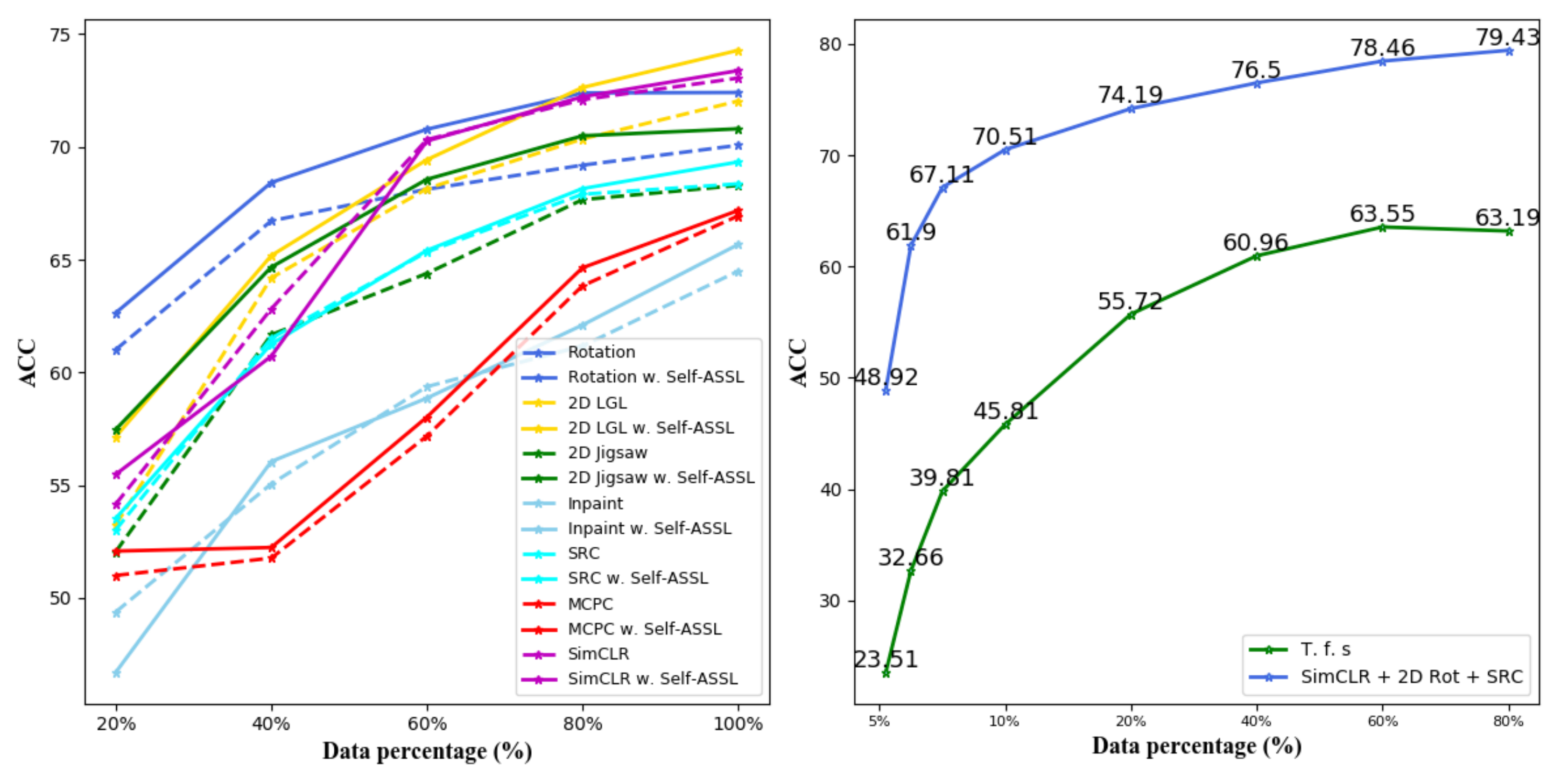}
\end{center}
\caption{The performances of networks trained with different amounts of labeled data on STL10 dataset. The left figure shows the results of various SSLs with/without Self-ASSL and the right shows improvement of MT-ASSL, respectively. Refer to \emph{supplementary material} for clear observation.}
\label{figure_data_number}\vspace{-5mm}
\end{figure}

\subsection{Performances on Extremely Limited Data}
We also conduct an experiment to evaluate the variation of accuracy achieved by different SSL approaches while training with different numbers of images from STL10. The evaluation results are shown in Fig.~\ref{figure_data_number}. The smallest amounts of training data are $20\%$ and $5\%$ of STL10 for MT-ASSL and Self-ASSL, respectively, which are extremely limited for network training.
The left figure of Fig.~\ref{figure_data_number} shows that our Self-ASSL consistently boosts the accuracy of SSL methods, especially 2D Jigsaw, with varying amounts of training data; while the right one demonstrates that our MT-ASSL can remarkably improve the model performance, compared to train-from-scratch, even with few labeled data (\emph{i.e.,} $5\%$).

\vspace{-10pt}
\section{Conclusion}
We propose two approaches for SSL aggregation training from a limited sample by exploiting the complementarity of multiple proxy tasks and self-complementarity to a single proxy task itself, respectively. 
We first propose an effective multi-task aggregative strategy to fuse multiple proxy tasks and extract the complementary features. Subsequently, a self-aggregative SSL, which is simple but effective, is implemented to aggregate self-complementary feature to boost the performance of a single SSL method. Our extensive experiments on two datasets with limited annotations show that the proposed aggregation strategies expose new insights for self-supervised learning and significantly improve the accuracy of learned features on the target tasks. 
Future work includes mining the feature complementarity among off-the-shelf networks for various vision tasks.

{\small

\bibliographystyle{ieee_fullname}
}

\end{document}